\documentclass[10pt, conference]{IEEEtran}
\IEEEoverridecommandlockouts

\usepackage{cite}
\usepackage{amsmath,amssymb,amsfonts}
\usepackage{algorithmic}
\usepackage{graphicx}
\usepackage{textcomp}
\usepackage{xcolor}

\usepackage{booktabs}
\usepackage{multirow}
\usepackage{url}
\newcommand{\ie}{{\emph{i.e.}}}
\newcommand{\eg}{{\emph{e.g.}}}
\newcommand{\parhead}[1]{\noindent\textbf{#1}}

\def\BibTeX{{\rm B\kern-.05em{\sc i\kern-.025em b}\kern-.08em
    T\kern-.1667em\lower.7ex\hbox{E}\kern-.125emX}}
\begin{document}

\title{High-Performance Few-Shot Segmentation with Foundation Models: An Empirical Study\\
\thanks{Corresponding Author: Lihe Zhang. Huchuan Lu is also with School of Future Technology, Dalian University of Technology.}
}

\author{\IEEEauthorblockN{Shijie Chang, Lihe Zhang, Huchuan Lu}
\IEEEauthorblockA{\textit{School of Information and Communication Engineering, Dalian University of Technology}, China \\
csj@mail.dlut.edu.cn, \{zhanglihe, lhchuan\}@dlut.edu.cn}
}

\maketitle

\begin{abstract}
Existing few-shot segmentation (FSS) methods mainly focus on designing novel support-query matching and self-matching mechanisms to exploit implicit knowledge in pre-trained backbones. However, the performance of these methods is often constrained by models pre-trained on classification tasks. The exploration of what types of pre-trained models can provide more beneficial implicit knowledge for FSS remains limited. In this paper, inspired by the representation consistency of foundational computer vision models, we develop a FSS framework based on foundation models. To be specific, we propose a simple approach to extract implicit knowledge from foundation models to construct coarse correspondence and introduce a lightweight decoder to refine coarse correspondence for fine-grained segmentation. We systematically summarize the performance of various foundation models on FSS and discover that the implicit knowledge within some of these models is more beneficial for FSS than models pre-trained on classification tasks. Extensive experiments on two widely used datasets demonstrate the effectiveness of our approach in leveraging the implicit knowledge of foundation models. Notably, the combination of DINOv2 and DFN exceeds previous state-of-the-art methods by 17.5\% on COCO-20$^i$. Code is available at \url{https://github.com/DUT-CSJ/FoundationFSS}.
\end{abstract}

\begin{IEEEkeywords}
Few-shot Segmentation, Foundation Model.
\end{IEEEkeywords}

\section{Introduction}

Significant progress in fully-supervised semantic segmentation is driven by large-scale pixel-level labeled datasets. It takes more than an hour to obtain pixel-level annotation of an image~\cite{cordts2016cityscapes}, which makes the labor cost of large-scale pixel-level annotated datasets expensive. And once the datasets are labeled pixel by pixel, it is difficult to add new categories to them. In response to the above challenge, few-shot segmentation (FSS), which aims to segment the corresponding object with unseen categories of the query image using only a few support image-mask pairs, has been proposed~\cite{shaban2017one}.

Based on different strategies for mining query-support information, existing FSS methods can be divided into two technical approaches: support-query matching mechanism~\cite{min2021hypercorrelation, hong2022cost, moon2023msi, peng2023hierarchical, shuai2023pgmanet, yang2023mianet} and self-support matching mechanism~\cite{fan2022ssp, wang2023focus}. 
The former mechanism matches support features with query features by designing novel prototypical learning or dense correlation modules. The latter mechanism obtains a coarse segmentation map through support-query matching and uses a self-support module to refine the segmentation map. The continuous emergence of novel matching methods improves the performance of FSS. However, existing FSS methods mainly focus on designing matching modules to mine the implicit knowledge in the frozen pre-trained backbone, neglecting the exploration on which types of pre-trained backbone's implicit knowledge are more beneficial for FSS. 
In this paper, we aim to find a combination of pre-trained backbones that is more advantageous for the FSS task.

Recently, various foundation models with powerful transfer learning and zero-shot capabilities have emerged. MAE~\cite{he2022mae} uses an intra-image self-supervised training paradigm and shows significant performance improvement when fine-tuned on downstream tasks. DINO~\cite{caron2021dinov1} and DINOv2~\cite{oquab2023dinov2} learn semantically invariant features with a discriminative self-supervised learning paradigm. Contrastive language-image pre-training (CLIP)~\cite{radford2021clip} can provide well-aligned textual and visual embeddings. The internal representations of text-to-image diffusion models~\cite{rombach2022sd} also demonstrate transfer capability to downstream tasks.
Many methods transfer the aforementioned foundation models to their respective downstream tasks, \eg, correspondence estimation and FSS, demonstrating superior performance.
DIFT~\cite{tang2023dift} extracts diffusion features to establish correspondences between images. 
SD-DINO~\cite{zhang2024sddino} proposes to fuse the features of Stable Diffusion (SD)~\cite{rombach2022sd} and DINOv2~\cite{oquab2023dinov2} for semantic and dense correspondence.
UniFSS~\cite{chang2024unifss} utilizes CLIP to build a universal vision-language framework to accomplish seven FSS tasks.
While the above prior works show that foundation models can be used for correspondence estimation and FSS, the performance of various foundation models in FSS has not been thoroughly analyzed. Which foundation models' implicit knowledge is more beneficial for FSS should be explored.

\begin{figure*}[t]
  \centering
  \includegraphics[width=0.9\linewidth]{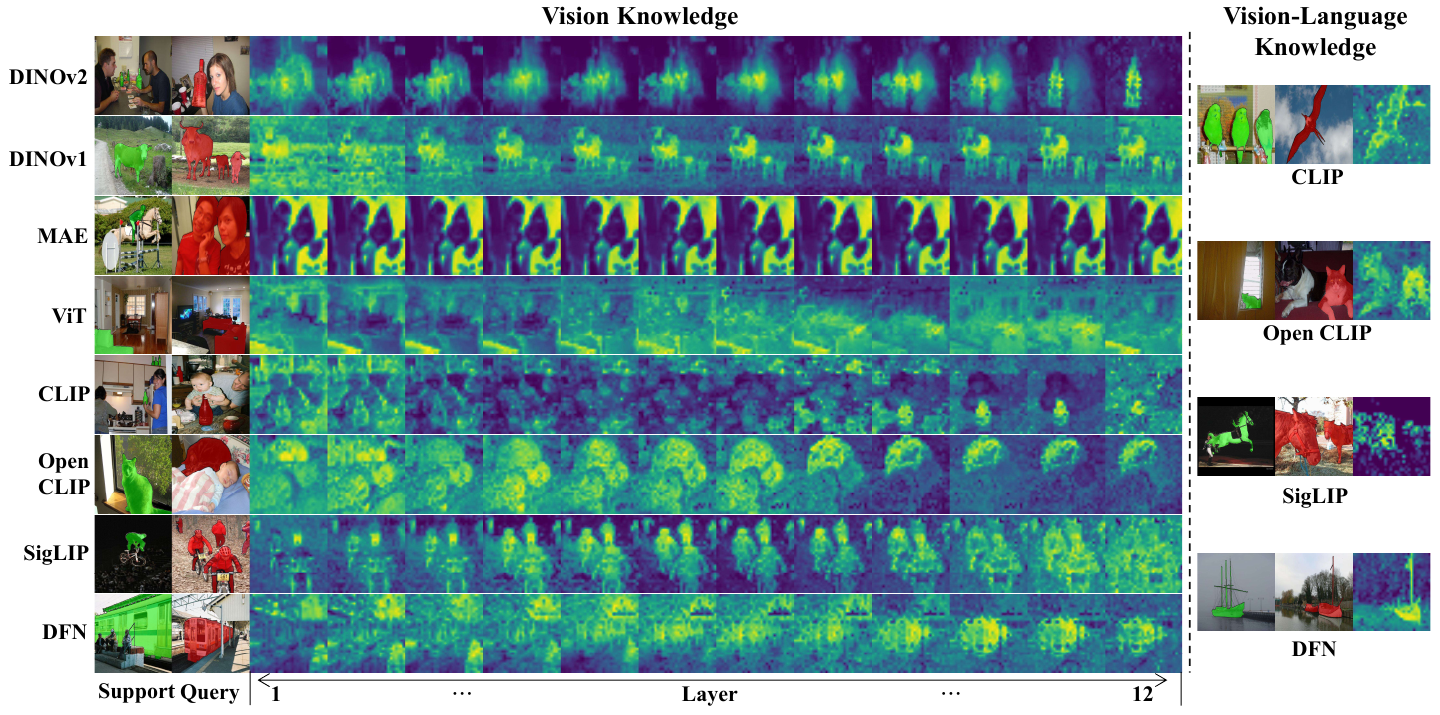}
  \caption{\textit{Left}: visualization of the vision knowledge of 7 foundation models and ViT pre-trained on the classification task. \textit{Right}: visualization of the vision-language knowledge of 4 foundation models. Due to space constraints, only one example is shown for each foundation model.}
  \label{fig:visualize}
\end{figure*}

To address the challenges above, we propose a simple framework that extracts implicit knowledge from foundation models to construct coarse correspondence and refine the coarse correspondence for fine-grained segmentation with a lightweight decoder. We systematically summarize the performance of various foundation models in FSS from both quantitative and qualitative perspectives, including DINO~\cite{caron2021dinov1}, DINOv2~\cite{oquab2023dinov2}, MAE~\cite{he2022mae}, CLIP~\cite{radford2021clip}, Open CLIP~\cite{ilharco_gabriel_2021_5143773,cherti2023reproducible,schuhmann2022laionb}, SigLIP~\cite{zhai2023siglip}, and DFN~\cite{fang2023dfn}. We explore the benefits of foundation models for FSS from a different perspective than previous work, which may inspire researchers to address FSS from additional perspectives.

Our contributions are summarized as follows:
\begin{itemize}
    \item We address FSS from a new perspective, focusing on which knowledge from pre-trained models benefits FSS, rather than designing new matching algorithms. We analyze the potential of several vision and vision-language foundation models for FSS.
    \item We propose a simple framework that leverages implicit knowledge from foundation models to establish coarse correspondences and a lightweight decoder to refine them for fine-grained segmentation.
    \item Experiments show that our method achieves a new state-of-the-art on PASCAL-5$^i$~\cite{shaban2017one} and COCO-20$^i$~\cite{lin2014microsoft} under both mask
    FSS and class-aware mask FSS settings. Ablation studies indicate that the combination of DINOv2 and DFN achieves the best performance, surpassing previous state-of-the-art methods by 17.5\% on COCO-20$^i$ in terms of mIoU. 
\end{itemize}

\section{Method}

\begin{figure}[t]
  \centering
  \includegraphics[width=\linewidth]{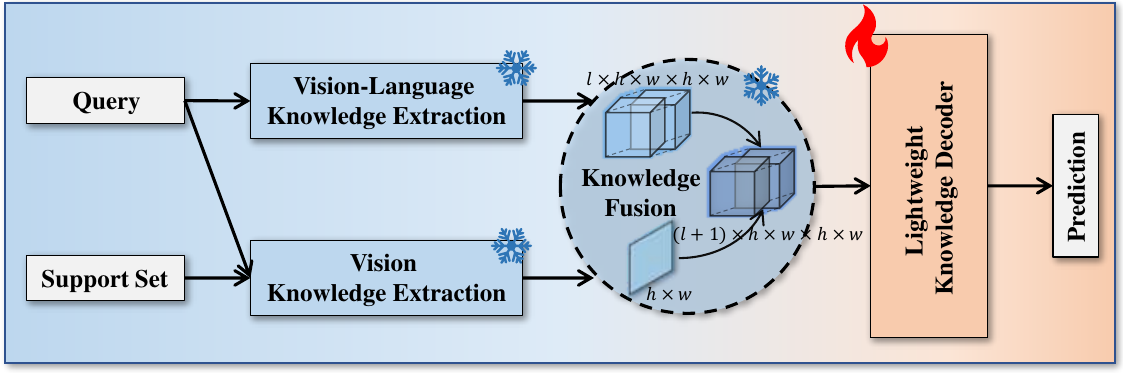}
  \caption{The architecture of our proposed framework.}
  \label{fig:framework}
\end{figure}

We first outline the problem setup of FSS-related tasks, then examine the properties of foundation models' implicit knowledge for FSS, and finally introduce a simple strategy and a lightweight decoder to harness this knowledge effectively.

\subsection{Problem Formulation}
\label{sec:task}

In FSS-related tasks, the dataset $\mathcal{D}$ is divided into disjoint $\mathcal{D}_\mathrm{base}$ with category set $\mathcal{C}_\mathrm{base}$ for training and $\mathcal{D}_\mathrm{novel}$ with unseen category set $\mathcal{C}_\mathrm{novel}$ for testing, \ie, $\mathcal{C}_\mathrm{base} \cap \mathcal{C}_\mathrm{novel} = \emptyset$. Episodes are utilized during both the training and testing phases.
Each episode includes a query set $\mathcal{Q} = \{I_q, M_q\}$ and support set $\mathcal{S} = \{\mathcal{S}_k\}_{k=1}^K$ with the same category $c$.
Following previous works, $\mathcal{S}_k$ comprises various types of support information, including image $I_s$, binary mask $M_s$, box $B_s$, and category label $T_s$.
With the support set, the model $f(\cdot,\theta)$ learns to map from the query image $I_q$ to the ground truth $M_q$. 
After training, the model performs episode testing without optimization.

\subsection{Analysis of the Implicit Knowledge in Foundation Models}
\label{sec:analysis}

We extract implicit knowledge from foundation models by leveraging their features. To ensure a fair comparison, we select the Vision Transformer Base (ViT-B) model of each foundation model as the feature extractor. 
7 foundation models are selected in our experiments, including DINOv2, DINOv1, MAE, CLIP, Open CLIP, SigLIP, and DFN.
DINOv2 and DINOv1 learn features by discriminative self-supervised learning. MAE learns feature representations by reconstructing masked images. CLIP, Open CLIP, SigLIP, and DFN are all language-image pre-training foundation models. Among them, Open CLIP uses a large-scale training set~\cite{schuhmann2022laionb}, SigLIP introduces sigmoid loss to scale up the batch size, and DFN designs a data filtering network to filter the large uncurated dataset. We also choose ViT~\cite{dosovitskiy2020vit} pre-trained on image classification tasks to compare with the foundation models.

\parhead{Vision Knowledge Extraction.} ViT-B consists of 12 transformer blocks. We extract features after each block, resulting in a total of 12 features.
We compute the cosine similarity between the query features and the masked support features to obtain a 4D similarity map. This process is formalized as
\begin{equation}
  C_{v}^{i} = \mathrm{ReLU}\left(\frac{\tilde{F}_q^{i} \cdot {\tilde{F}_s^{i\top}}}{\left \| \tilde{F}_q^{i} \right \| \cdot \left \| \tilde{F}_s^{i} \right \|}\right),
  \label{eq:corr_vv}
\end{equation}
where $C_{v}^i \in \mathbb{R}^{h \times w \times h \times w}$. $\tilde{F}_q^{i} \in \mathbb{R}^{h w \times c}$ and $\tilde{F}_s^{i} \in \mathbb{R}^{h w \times c}$ are query and masked support features after reshaping. By averaging over the last two dimensions of $C_{v}^i$, we obtain a 2D activation map for the query image. 
Fig.~\ref{fig:visualize} shows the visualization of the vision knowledge extracted from different models, including several foundation models and the ViT pre-trained on the classification task. 
The visualization indicates the following results: 1) the last layer of DINOv2 accurately locates the target, 2) layers 2-12 of DINOv1 roughly locate the target but contain background noise, 3) MAE and ViT fail to locate the target, 4) The last four layers of CLIP, Open CLIP, and DFN can roughly locate the target, 4) the middle layers of SigLIP provide a rough localization of the target. Among them, DINOv2 shows the best qualitative performance.

\parhead{Vision-language Knowledge Extraction.} Vision-language pre-trained foundation models learn a unified modality representation. The implicit knowledge in these models can also be activated by textual information. We extract vision-language implicit knowledge $C_{t} \in \mathbb{R}^{h \times w}$ by computing the cosine similarity between the features from the visual and textual encoder~\cite{zhou2022maskclip}.
The visualization of visual-language knowledge is on the right of Fig.~\ref{fig:visualize}.
DFN can relatively accurately locate the target, while SigLIP contains the least background noise. Compared to the other three models, DFN exhibits the best qualitative performance.

The above observations of vision and vision-language knowledge suggest that combining DINOv2 and DFN may yield the best location of targets.  In Sec.~\ref{sec:ablation}, we conduct a detailed study on different combinations of foundation models, proving that combining DINOv2 and DFN achieves the best performance.

\subsection{Decoding Implicit Knowledge for FSS}
\label{sec:decoder}

Following UniFSS, we present our proposed method in the context of class-aware mask FSS under the 1-shot setting, \ie, $\mathcal{S} = \{I_s, M_s, T_s\}$ and $\mathcal{Q} = \{I_q, M_q\}$. As for $K$-Shot inference, the model performs 1-shot inference $K$ times to generate $K$ prediction maps, which are then combined through voting to produce the final prediction. The overview of the proposed method can be found in Fig.~\ref{fig:framework}.

\parhead{Knowledge Fusion.} Sec.~\ref{sec:analysis} introduces how to extract implicit knowledge from foundation models such as DINOv2 and CLIP. We obtain 4D vision activation map set $\{\mathcal{C}_i\}_{k=m}^{12}$ and 2D textual activation map $C_{t}$.
There are two methods of integrating $\{\mathcal{C}_i\}_{k=m}^{12}$ and $C_{t}$, \ie, early fusion and late fusion. In our experiments, we found early fusion performs better. 
We broadcast the dimensions of $C_{t}$ to match those of $C_{i}$ and concatenate $C_{t}$ with $\{\mathcal{C}_i\}_{k=m}^{12}$ to obtain the fused knowledge $C_{f} \in \mathbb{R}^{l \times h \times w \times h \times w}, l=12-m+1$, $m$ is a hyper-parameter that controls which layers are used for knowledge extraction.
Further experimental details can be found in Sec.~\ref{sec:ablation}.

\parhead{Lightweight Knowledge Decoder.} After the knowledge fusion, a lightweight Knowledge decoder is applied to $C_{f}$ to obtain an accurate prediction map.
We first encode $C_{f}$ into a high-dimension representation $C_{f}^{'} \in \mathbb{R}^{d \times h \times w \times h^{'} \times w^{'}}$ using center-pivot 4D convolution~\cite{min2021hypercorrelation}, ReLU, and group normalization (GN).
Previous works use complex modules such as 4D Swin Transformer~\cite{hong2022cost} and 4D Deformable Transformer~\cite{xiong2022doubly} to refine $C_{f}^{'}$.
However, these approaches introduce redundant parameters. Notably, the implicit knowledge of foundation models can achieve high performance with lightweight decoder modules.
To achieve this, we designed a depth-wise separable 4D convolution module (DSCM) that includes depth-wise 4D convolution, point-wise 4D convolution, activation function, and GN. Specifically, we decompose the center-pivot 4D convolution into depth-wise and point-wise 4D convolutions. DSCM is formalized as
\begin{equation}
  \begin{aligned}
  C_{f}^{n+1} = \mathrm{DSCM}(C_{f}^{n}) + C_{f}^{n}
  \end{aligned}
  \label{eq:dscm}
\end{equation}
where each $\mathrm{DSCM}$ repeats the following process three times,
\begin{equation}
  \begin{aligned}
  C_{f}^{n'} =& \mathrm{ReLU}(\mathrm{ReLU} (\mathrm{PW4DConv}\\&(\mathrm{DW4DConv}(C_{f}^{n})))).
  \end{aligned}
  \label{eq:dscm_detail}
\end{equation}
Finally, the 4D map $C_{DSCM} \in \mathbb{R}^{d \times h \times w \times h^{'} \times w^{'}}$ obtained after two layers of DSCM is averaged over $h^{'} \times w^{'}$ dimension to obtain a 2D feature map $C \in \mathbb{R}^{d \times h \times w}$. $C$ is then upsampled and refined by stacked convolutions~\cite{he2016deep} to get the prediction map.

\section{Experiments}



\begin{table*}[!t]
  \centering
  \begin{minipage}[t]{0.74\linewidth} 
    \centering
    \caption{Performance comparison on PASCAL-5$^{i}$~\cite{shaban2017one} and COCO-20$^i$~\cite{lin2014microsoft}.
  Numbers in \textbf{bold} indicate the best performance.
  VB: vision backbone, VLB: vision-language backbone, F$0$-F$3$: Fold0-Fold3.}
    \resizebox{1\linewidth}{!}{ 
      \begin{tabular}{c|cc|ccccc|ccccc|c}
  \toprule[2pt]
  & & & \multicolumn{5}{c|}{1-shot} & \multicolumn{5}{c|}{5-shot} & Learnable  \\
  \multirow{-2}{*}{Methods} & VB & VLB & F$0$ & F$1$ & F$2$ & F$3$ & mIoU & F$0$ & F$1$ & F$2$ & F$3$ & mIoU & params \\
  \midrule[1pt]
  \multicolumn{14}{c}{\textbf{Mask FSS on PASCAL-5$^i$}}   \\
  \midrule[1pt]
  MSI~\cite{moon2023msi} & ResNet101 & - & 73.1 & 73.9 & 64.7 & 68.8 & 70.1 & 73.6 & 76.1 & 68.0 & 71.3 & 72.2 & - \\
  UniFSS~\cite{chang2024unifss} & CLIP & - & 72.7 & 75.6 & 63.7 & 66.9 & 69.7 & 75.4 & 77.1 & 67.9 & 69.9 & 72.6 & 8.1M   \\
  Ours  & DINOv2 & - & \textbf{76.5} & \textbf{81.3} & \textbf{72.1} & \textbf{77.4} & \textbf{76.8} & \textbf{79.5} & \textbf{84.8} & \textbf{75.8} & \textbf{82.5} & \textbf{80.7} & 0.6M \\
  \midrule[1pt]
  \multicolumn{14}{c}{\textbf{Class-aware Mask FSS on PASCAL-5$^i$}}   \\
  \midrule[1pt]
  PGMANet~\cite{shuai2023pgmanet} & CLIP & CLIP & 74.0 & 81.9 & 66.8 & 73.7 & 74.1 & 74.5 & 82.2 & 67.2 & 74.4 & 74.6 & 2.7M \\
  PI-CLIP~\cite{wang2024rethinking} & ResNet50 & CLIP & 76.4 & \textbf{83.5} & 74.7 & 72.8 & 76.8 & 76.7 & 83.8 & 75.2 & 73.2 & 77.2 & 4.2M \\
  UniFSS~\cite{chang2024unifss} & CLIP & CLIP & 75.0 & 79.6 & 74.7 & 76.4 & 76.4 & 75.5 & 79.9 & 75.9 & 77.5 & 77.2 & 8.1M \\
  Ours    & DINOv2 & DFN & \textbf{78.1} & 83.2 & \textbf{76.9} & \textbf{80.6} & \textbf{79.7} & \textbf{79.4} & \textbf{84.6} & \textbf{78.7} & \textbf{83.6} & \textbf{81.6} & 0.6M \\
  \midrule[1pt]
  \multicolumn{14}{c}{\textbf{Mask FSS on COCO-20$^i$}}   \\
  \midrule[1pt]
  MSI~\cite{moon2023msi} & ResNet101 & - & 44.8 & 54.2 & 52.3 & 48.0 & 49.8 & 49.3 & 58.0 & 56.1 & 52.7 & 54.0 & - \\
  UniFSS~\cite{chang2024unifss} & CLIP & - & 46.5 & 53.0 & 48.0 & 48.2 & 48.9 & 50.3 & 59.5 & 54.4 & 52.0 & 54.1 & 8.1M \\
  Ours    & DINOv2 & - & \textbf{56.0} & \textbf{61.3} & \textbf{57.9} & \textbf{58.8} & \textbf{58.5} & \textbf{61.4} & \textbf{69.4} & \textbf{65.9} & \textbf{64.9} & \textbf{65.4} & 0.6M \\
  \midrule[1pt]
  \multicolumn{14}{c}{\textbf{Class-aware Mask FSS on COCO-20$^i$}}   \\
  \midrule[1pt]
  PGMANet~\cite{shuai2023pgmanet} & CLIP & CLIP & 55.2 & 62.7 & 60.3 & 59.4 & 59.4 & 55.9 & 65.9 & 63.4 & 61.9 & 61.8 & 2.7M \\
  PI-CLIP~\cite{wang2024rethinking} & ResNet50 & CLIP & 49.3 & \textbf{65.7} & 55.8 & 56.3 & 56.8 & 56.4 & 66.2 & 55.9 & 58.0 & 59.1 & 4.2M \\
  UniFSS~\cite{chang2024unifss} & CLIP & CLIP & 51.2 & 61.8 & 58.0 & 55.6 & 56.7 & 53.1 & 62.4 & 59.2 & 56.8 & 57.9 & 8.1M \\
  Ours    & DINOv2 & DFN & \textbf{59.1} & 64.5 & \textbf{62.5} & \textbf{62.7} & \textbf{62.2} & \textbf{62.8} & \textbf{71.6} & \textbf{65.8} & \textbf{65.9} & \textbf{66.5} & 0.6M \\
  \bottomrule[2pt]
\end{tabular}

    }
    \label{tab:sota}
  \end{minipage}
  \hfill
  \begin{minipage}[t]{0.255\linewidth} 
    \centering
    \caption{Ablation study on fold$0$ of PASCAL-5$^{i}$~\cite{shaban2017one}. VB: vision backbone, VLB: vision-language backbone, *: only vision knowledge of the last layer is used.}
    \resizebox{1\linewidth}{!}{ 
\begin{tabular}{c|cc}
  \toprule[2pt]
   Model & 1-shot  & 1-shot \\
  \midrule[1pt]
  \multicolumn{3}{c}{\textbf{Ablation on Knowledge Fusion}}   \\
  \midrule[1pt]
  Late Fusion & 76.3 & 77.6 \\
  Ours(Early Fusion) & 78.1 & 79.4 \\
  \midrule[1pt]
  \multicolumn{3}{c}{\textbf{Ablation on VLB}}   \\
  \midrule[1pt]
  DINOv2 + CLIP & 76.1 & 77.3 \\
  DINOv2 + Open CLIP & 75.1 & 76.5 \\
  DINOv2 + SigLIP & 76.6 & 77.6 \\
  DINOv2 + DFN & 78.1 & 79.4 \\
  \midrule[1pt]
  \multicolumn{3}{c}{\textbf{Ablation on VLB}}   \\
  \midrule[1pt]
  DINOv1 + DFN & 72.0 & 73.5 \\
  MAE + DFN & 59.0 & 59.7 \\
  ViT + DFN & 71.2 & 71.9 \\
  SigLIP + DFN & 68.4 & 68.6 \\
  DFN + DFN & 64.4 & 69.2 \\
  DINOv2* + DFN & 74.2 & 75.9 \\
  DINOv2 + DFN & 78.1 & 79.4 \\
  \bottomrule[2pt]
\end{tabular}
    }
    \label{tab:ablation}
  \end{minipage}
\end{table*}

\subsection{Experimental Setup}

\parhead{Datasets.} We conduct experiments on two common FSS datasets, \ie,  PASCAL-5$^i$~\cite{shaban2017one} and COCO-20$^i$~\cite{lin2014microsoft}. PASCAL-5$^i$ comprises PASCAL VOC 2012~\cite{everingham2010pascal} along with additional mask annotations~\cite{hariharan2014simultaneous}. It consists of 20 classes, divided into 4 folds for cross-validation. COCO-20$^i$ is generated from MS-COCO~\cite{lin2014microsoft}. Its 80 classes are split into 4 folds, each containing 20 classes.

\parhead{Evaluation metric.} Building upon prior works~\cite{moon2023msi}, we use mean intersection over union (mIoU) as our evaluation metrics. mIoU computes the average IoU across all classes within each fold. 

\parhead{Implementation Details.} All experiments are implemented in PyTorch~\cite{paszke2019pytorch} and optimized using Adam with a fixed learning rate of 0.001. The spatial resolutions of features are set to $30 \times 30$ throughout all experiments. 
During training, the parameters are optimized by cross-entropy loss. All experiments are conducted on a single RTX 3090 GPU with 24G memory.
In the class-aware mask FSS setting, we choose DINOv2 as the visual backbone to extract vision knowledge, while DFN is used as the vision-language backbone to extract vision-language knowledge. As for the mask FSS setting, the vision-language backbone is removed, leaving only DINOv2 as the visual backbone. We set hyper-parameter $m=0$ in our experiments, \ie, vision knowledge from all layers is utilized.

\subsection{Comparison with State-of-the-Art Methods}

We evaluate our proposed method on PASCAL-5$^{i}$~\cite{shaban2017one} and COCO-20$^i$~\cite{lin2014microsoft} and compare the results with previous state-of-the-art methods. All results are shown in Tab.~\ref{tab:sota}.

\parhead{Mask FSS.} Mask FSS is the most common setting where the model takes image-mask pairs as support set.
Our approach significantly outperforms previous methods with fewer learnable parameters. Our approach achieves a relative mIoU improvement of 9.6\% and 17.5\% under the 1-shot setting for PASCAL-5$^{i}$ and COCO-20$^i$, respectively. Increasing from $1$-shot to $5$-shot, the performance improvement of our method is significantly better than the previous state-of-the-art methods. This indicates that DINOv2 possesses implicit knowledge beneficial to FSS and our strategy effectively extracts and refines it.

\parhead{Class-aware Mask FSS.} In this setting, category labels are provided in the support set. 
With the help of the vision-language knowledge extracted from the vision-language backbone DFN, our method achieves a 6.3\% relative mIoU improvement compared to using vision knowledge only on COCO-20$^i$. From 1-shot to 5-shot, PGMANet only increases the relative mIoU by 4.0\% on COCO-20$^i$, while our method increases the relative mIoU by 6.9\%. This indicates that our approach exhibits more potential for performance improvement as the number of images in the support set increases.

\subsection{Ablation Study}
\label{sec:ablation}

We conducted ablation studies from three aspects: knowledge fusion methods, vision-language foundation models, and vision backbones.

\parhead{Knowledge Fusion.} Early fusion is the strategy proposed in Sec.~\ref{sec:decoder}. The late fusion strategy refers to using DSCM and 2D convolution to decode vision knowledge and vision-language knowledge respectively, followed by convolutions to fuse the two kinds of decoded knowledge. The results show that early fusion of implicit knowledge achieves better performance.

\parhead{Vision-Language Foundation Models.} Vision-language foundation models can provide rich implicit knowledge to locate targets. Among the four models, DFN achieved the best qualitative and quantitative performance. SigLIP obtains the second place. This indicates that compared to the original CLIP, data filtering and scaling up the training batch size can facilitate the model in learning better implicit representations.

\parhead{Vision Backbones.} We evaluate the performance of different vision backbones when paired with DFN. DINOv2 significantly outperforms other vision backbones in terms of mIoU. Discriminative self-supervised pre-trained models, DINOv1 and DINOv2, outperform ViT pre-trained on the classification task. However, using MAE, SigLIP, and DFN as vision backbone performs below ViT. Additionally, experiments on the number of layers used to extract vision knowledge show that using all layers outperforms using only the last layer.
Experiments indicate that certain foundation models, such as DINOv2 and DINOv1, have significant potential in FSS.

\section{Conclusion}

In this paper, we address FSS from a new perspective, focusing on which knowledge from pre-trained models facilitates FSS.
To address this, we propose a simple strategy to extract implicit knowledge from foundation models and introduce a lightweight decoder to obtain fine-grained segmentation.
Build upon this, we systematically summarize the performance of multiple foundation models in FSS both qualitatively and quantitatively.  We find that the implicit knowledge of DINOv2 and DFN is more beneficial for FSS.
We hope our empirical study can provide new perspectives for FSS.

\ifCLASSOPTIONcaptionsoff
  \newpage
\fi
\bibliographystyle{IEEEtran}
\bibliography{IEEEabrv,refs}


\end{document}